\def\year{2018}\relax
\documentclass[letterpaper]{article} 
\usepackage{aaai18}  
\usepackage{times}  
\usepackage{helvet}  
\usepackage{courier}  
\usepackage{url}  
\usepackage{graphicx}  
\usepackage[utf8]{inputenc} 
\usepackage[T1]{fontenc}    
\usepackage{url}            
\usepackage{booktabs}       
\usepackage{amsfonts}       
\usepackage{nicefrac}       
\usepackage{graphicx,amsmath,amssymb,bm,dsfont}      
\usepackage{mysubfigure,enumitem}      
\usepackage{mathrsfs}      
\usepackage{amsthm}      
\usepackage[ruled,linesnumbered,noresetcount]{algorithm2e}      

\newcommand{\newcomment}[1]{}
\newcommand{\scr}[1]{\ensuremath{\mathcal{#1}}}

\newcommand{\B}[1]{\ensuremath{\mathbf{#1}}}
\newtheorem{proposition}{Proposition}
\title{Predicting vehicular travel times  by modeling heterogeneous influences between arterial roads}
\author{Avinash Achar \and Venkatesh Sarangan \and Rohith R\\
Innovation Labs, Tata Consultancy Services,\\ 
Chennai.\\
\And 
Anand Sivasubramaniam\\ 
Dept. of Comp. Sci \& Eng.,\\  
Pennsylvania State University, USA.}
 
\begin{document}
%
\maketitle
\begin{abstract}
Predicting travel 
times of vehicles in urban settings  is a useful and tangible quantity of interest
in the context of intelligent transportation systems.
We address the problem of travel time prediction in arterial roads 
using data sampled from probe vehicles.  
There is only a limited
literature on methods using  data input from probe vehicles. 
The spatio-temporal dependencies captured by existing data driven approaches are either too detailed or very simplistic.  
We strike a balance of the existing data driven approaches to account for varying degrees of  
influence a given road  may experience from its neighbors, while controlling the number of parameters to be learnt.
Specifically, we use a NoisyOR conditional probability 
distribution (CPD) in conjunction with a dynamic bayesian network (DBN)
to model  state transitions of various roads. We propose 
an efficient algorithm to learn model parameters.
We propose an algorithm for predicting travel times on trips of arbitrary durations.
Using synthetic and real world data traces we demonstrate the superior performance of the 
proposed method under different traffic conditions.
\end{abstract}

\section{Introduction}
\label{sec:introduction}
{\bf Travel-time prediction:} 
Advances in affordable technologies for sensing and communication have 
allowed us to gather data about large distributed infrastructures
such as road networks in real-time. The collected data is digested to 
generate information that is useful for the end users  (namely commuters) as well
as the road network administrators. 
From the commuters' perspective, travel time is perhaps the 
most useful information. Predicting travel time along various
routes in advance with good accuracy allows commuters to 
plan their trips appropriately by identifying and avoiding congested 
roads. This can also aid traffic administrators to make crucial real-time decisions
for mitigating prospective congestions, design infrastructure changes for better mobility and so on.
Crowd-sourcing based applications such as Google Maps allow 
commuters to predict their travel times along multiple routes. 
While the prediction accuracy of such applications
is reasonable in many instances, they may not be
helpful for all vehicles. 
In certain countries, vehicles  such as small 
commercial trucks are restricted to specific lanes with  their own  different (often lower) speed limit. 
 Hence, the travel times and congestions seen by such vehicles could be different from the
(possibly average) values that are predicted from 
crowd sourced applications. 
In such cases, customized 
travel-time prediction techniques are necessary.

{\bf Types of prediction models:} 
Travel time 
 prediction models can be broadly categorized into two 
types:  traffic flow based and data-driven \cite{Mori15}.
The traffic flow models attempt to capture the physics of the traffic in varied levels of detail. They however suffer from important issues like  need for calibration,
being computationally expensive and rendering inaccurate  predictions.

{\bf Data-driven models:}
The data driven models 
typically use 
statistical models which model traffic behavior to an extent just enough for the required 
prediction  at hand. They rely on real world data feeds for 
learning the parameters of the employed  statistical model. 
A  variety of data driven 
techniques to predict travel time  have been
proposed in the literature.  Researchers have proposed techniques based
on linear regression \cite{Kwon00,nikovski05}, time-series models \cite{Ishak03,Vanajakshi09},
neural networks \cite{li11}, 
regression trees \cite{Kwon00,nikovski05} and bayesian networks \cite{Hunter09} to name a few. 

Prediction in a freeway context (flow, travel time etc.) has been typically better studied compared to urban or arterial roads.
This is because freeways  are relatively well instrumented with sensors like  loop detectors, 
AVI detectors and cameras. On the other hand,  
urban/arterial roads have been relatively less studied
owing to complexities involved in handling traffic lights and intersections.
 Nevertheless, spread of GPS fixtures in vehicles/smart phones has rendered  probe vehicle data a reasonable data source for arterial traffic \cite{Liu2013,Aslam2012}. Recently, 
DBN  based approaches  have been proposed to predict travel time on arterial roads based on sparse probe vehicle data \cite{herring10,hofleitner12,hofleitner12a}. 
Under real world traffic conditions, these various DBN techniques have been shown to significantly outperform other simpler methods such as time-series models.

{\bf Gaps and contributions:} 
Current DBN based 
modeling approaches of congestion dependencies in road networks are either too meticulous to be used in large networks or too simplistic to be accurate. The
modeling assumption in \cite{herring10}, albeit quite general, leads to an exponential number of model parameters.
On the other hand,
the model proposed in \cite{hofleitner12} even though  has a tractable number of parameters, assumes that the state of congestion in  a    
given road is influenced {\it equally} by the state of congestion of all its neighbors, which can be pretty restrictive.
In reality, different neighbors will exert different degrees of influence on a given 
road  --  for instance, the state of a downstream road which receives bulk of the 
traffic from an upstream road will exert a higher influence on the 
congestion state of the upstream road than other neighbors.  
In this paper, we propose a novel DBN based approach that models the individual influence of different neighbors while remaining computationally tractable.
 Our specific {\bf contributions} include:
\begin{itemize}
\item We propose to use a `NoisyOR' CPD for modeling the varying degrees of influence of different neighbors of a
road. The degree of influence is offered as a separate parameter for each neighboring link.
It also keeps the number of parameters to be learnt linear in the number of neighboring links. 

\item  We develop a novel   Expectation-Maximization (EM) based algorithm to learn the DBN parameters under the above NoisyOR CPD. 

\item We propose a new algorithm for predicting travel time of a generic trip that can span 
an arbitrary duration.  Existing works can only handle trips that get completed within one DBN time step only. 

\item We test usefulness of our approach on both synthetic data and real-world probe vehicle data obtained
from (i)city of Porto, Portugal and (ii)San Fransisco. On synthetic data, relative absolute prediction error  can reduce by as much as  $70\%$  under the proposed method in the
worst case.
On real world data traces from Porto and San Fransisco, the proposed approach performs up to $14.6\%$  and $16.8\%$ better respectively than existing approaches in the worst case.
\end{itemize}
We note here that the proposed DBN with NoisyOR CPD transitions can be used in other domains as well, such as BioInformatics (more details in Section
\ref{sec:Conclusions}).
Therefore, the proposed method has a wider reach than the specific transportation application discussed in detail in this paper.

\section{Related work}
\label{sec:RelWork}
Research based on probe vehicle data has been steadily on the rise of late given the wide spread of GPS based sensing. 
Proble data has been utilized for various tasks like traffic volume and hot-spot estimation \cite{Aslam2012}, adaptive routing \cite{Liu2013},  estimation and
prediction of travel time and so on.  Travel time {\em estimation}\footnote{Travel time estimation 
is the task of computing travel times of trips or trajectories that have already been completed, while prediction involves trips that start in the future.} is another
(well studied) important task  useful in particular for traffic managers. Since our focus in this paper is on prediction alone and since most of the travel estimation methods 
do not have predictive abilities, we do not elaborate
on this further here. Please refer to App.~\ref{sec:TTE} for a summary.

Literature on arterial travel time prediction  using probe vehicles has been relatively sparse.  We focus on DBN approaches which explicitly model the congestion state at
each link. Among such DBN approaches,
a hybrid approach that combines traffic flow theory and DBNs is proposed in 
\cite{hofleitner12a}. It captures flow conservation, uses traffic theory inspired travel time distributions, its state variables are no more binary 
but the queue length built at each link. However, as discussed in \cite{HofleitnerPhD} some of the model assumptions made in \cite{hofleitner12a} like uniform arrivals are 
too strong and have limitations compared with physical reality.  \cite{HofleitnerPhD} goes on to vouch for a relatively more data-driven approach as proposed in \cite{hofleitner12}.
Our proposed work is closely related to  \cite{herring10} and \cite{hofleitner12}. 
In fact, our proposed approach tries to incorporate the best of these  approaches while
circumventing their drawbacks.
The approach in \cite{herring10} 
leads to an exponential number of parameters that have to be learnt and hence suffers from severe overfitting. 

\section{DBN Model}
\label{sec:Model}

{\bf Input Data:} Probe vehicles are a sample of vehicles plying around the road network providing periodic information about 
their location, speed, path etc. Such vehicles act as a data source for observing the network's condition. 
Such historical
data  is used for learning the underlying DBN model parameters.  The learnt parameters  
along with current real-time probe data are used to perform short-term travel time predictions across the network. 
Real time is discretized into time bins (epochs or steps) of uniform size $\Delta$. At each time 
epoch $t$, we have a set of probe vehicle trajectory measurements. Each trajectory is specified by 
its start and end ($x_s$ and $x_e$)  which comes from successively sampled location co-ordinates, and  
sequence of links traversed in moving from
$x_s$ to $x_e$. 
The data input to the algorithm is the set of all such trajectories collected over multiple time epochs. Notationally, for the 
$k^{th}$ vehicle's trajectory
at time step $t$, ${x}_{s,t}^k$ and  ${x}_{e,t}^k$ are its start and end locations, and $L_t(k)$ is the sequence of links traversed. If $N^v_t$
denotes the  
number of active vehicles at time step $t$, then the index $k$ at time step $t$ can vary from $1$ to $N^v_t$. Note that 
$N^v_t$ is a function of $t$ in general.
In order to filter GPS noise and  obtain path information,  map matching and path-inference algorithms  \cite{Hunter11} can be used.
For ease of reference, notation used in this paper is summarized in App.~\ref{sec:table} .
\subsection{DBN Structure }

\begin{figure}[h]
\center
\subfigure[Two time-slice bayesian net (2TBN) structure.]{
\includegraphics[width=3.0in,height=1.30in]{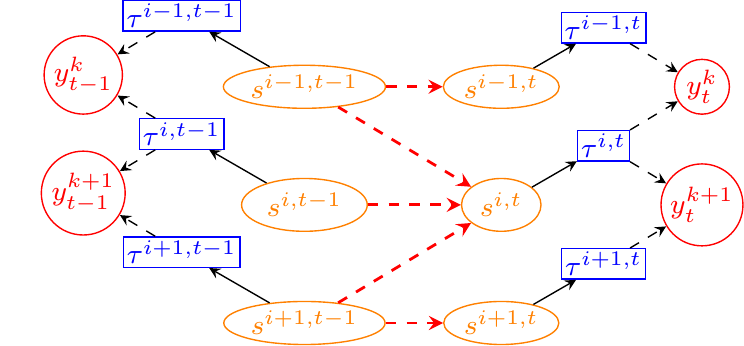}
\label{fig:TTS}
}
\subfigure[NoisyOR: Proposed transition CPD model.]{
\includegraphics[width=3.00in,height=1.30in]{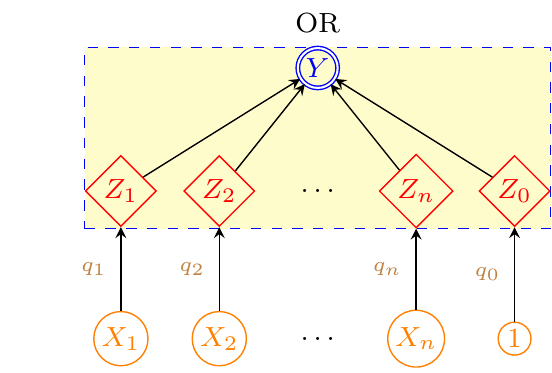}
\label{fig:NoisyOR}
}
\caption{DBN structure and the proposed Transition CPD.}
\label{fig:DBN}
\end{figure}

Fig.~\ref{fig:TTS} shows the DBN structure  \cite{herring10,hofleitner12} that we use in this paper to capture spatio-temporal dependencies between links of the network.
The arterial traffic is modeled as a discrete-time dynamical system.
At each time step $t$, a link $i \in \mathcal{I}$ in the network is assumed to be in one of two states namely, congested ($1$) or 
uncongested ($0$). $s^{i,t}$ denotes this 
state of congestion at link $i$  and time $t$. Note that these are hidden state variables as far as the model is  concerned.
We denote by $\pi_i$, the set of roads that are adjacent (both upstream and downstream) to road $i$ including itself. The adjacency structure of 
the road network is  utilized to obtain the transition structure  of the DBN from time step $t$ to time $t+1$. 
Specifically, the state of a link $i$ at time $t+1$ is assumed to be a function of the state of all its neighbors
$\pi_i$ at time $t$. In the DBN structure, this implies that the node corresponding
to the link $i$ at time $t+1$ will have incoming edges from nodes in $\pi_i$ at time $t$. 
\newcomment{
Consequently, given the state of neighbors   at time $t$, 
the state of link $i$ at time $t+1$ is independent of the state of the non-neighboring links at $t$  and 
is independent  of the state of all links before $t$.            
}

We assume the travel time on a link to be a 
random variable whose distribution depends on the state of the respective link. The traversal time on a trajectory is a  sum of  random variables, each representing
the travel time of a (complete or partial) link of the trajectory. From the structure of the DBN (fig.~\ref{fig:TTS}), given the 
state information of
the underlying links, these link travel times ($\tau^{i,t}$, denoted as rectangles in fig.~\ref{fig:TTS}) are {\em independent}.  
Hence the conditional travel time on a path is a sum of independent random variables.
In general,  the first and last links in the set $L_j(k)$ get partially traversed. 
In such cases, one can obtain the partial link travel times by  scaling (linear or non-linear \cite{herring10,hofleitner12}) the complete link travel time  as per the distance. 
In this paper, we use 
linear scaling.

\subsection{Conditional probability distributions on DBN }
\label{sec:keydiff}
\newcomment{
We have two distribution types: (a){\it Observed CPD}:  governs 
observations given the current state of the links, (b){\it Transition
CPD}: governs link states at current time  given the  link states at previous time.
}

{\bf Observation CPD:} 
Travel time distribution on a link $i$ given its state $s$, is assumed to be normally distributed with parameters
$\mu^{i,s}$ and $\sigma^{i,s}$.  We compactly refer to these observation parameters as ($\bm{\mu}, \bm{\sigma}$). The  travel time measurement 
from the $k^{th}$ vehicle at time epoch $t$, $y_{t,k}$ (denoted by circles in fig.~\ref{fig:TTS}) 
is specified by the set of links
traversed, $L_t(k)$, and position of the start and end co-ordinates on the first and last links (namely $x_{s,t}^k$ and $x_{e,t}^k$ respectively).
$f(y^k_t|s^{L_t(k),t},x_{s,t}^k, x_{e,t}^k)$ denotes the conditional distribution of a travel-time measurement, conditioned on the links traversed and start-end
positions. Given state information of links along a path, owing to normality and conditional independence of these travel times, travel time on any path is also 
normally 
distributed. The associated mean and variances are sum of mean and variances of 
individual link (possibly scaled) travel times.

{\bf Existing Transition CPDs:} 
Let $A(\bm{\eta}^{i,t-1},s^{i,t})$ be the CPD that models 
influence exerted on road $i$'s  state at time $t$ by $\bm{\eta}^{i,t-1}$, 
the states of its neighbors at time $t-1$. If this factor is  a general
tabular CPD  as proposed in \cite{herring10}, then number of parameters  grows exponentially with 
number of neighbors. 

To circumvent this,  \cite{hofleitner12}
chooses a CPD whose number of parameters is
linear in the number of neighbors.
Instead of a separate bernoulli distribution for each realization of $\bm{\eta}^{i,t-1}$, it looks at the
number of congested (or saturated) neighbors in the road network or parents in the DBN.
Hence we refer to this  as SatPat CPD in the rest of this paper.
If $a_{i,j}$ denotes the chance of congestion at the $i^{th}$ link 
given exactly  $j$ of its neighbors are congested at the previous time instant, then 
\vspace{-0.090in}
\begin{equation}
A(\bm{\eta}^{i,t-1}, s^{i,t}) =  \prod_{j=0}^{|\pi_i|} (a_{i,j})^{N^{i,t-1}_j s^{i,t}} (b_{i,j})^{N^{i,t-1}_j(1-s^{i,t})}
\label{eq:TransProbFactorSatPat}
\end{equation}
where $b_{i,j} = 1 - a_{i,j}$, and $N^{i,t-1}_j$ is an indicator random variable which is $1$ only when exactly $j$ of link $i$'s neighbors are congested. As mentioned earlier, this CPD has a few shortcomings:
\begin{itemize}[itemsep=-0.0cm,topsep=0ex,leftmargin=*]
\item It assumes all neighbors of a road have identical influence on a road's state.
In particular it assumes an identical congestion probability (namely $a_{i,1}$) at $i$  at time $t$, given exactly one of its neighbors is
congested at $t-1$.
This is irrespective of which of  $i$'s  neighbors is congested at $t-1$. 
\item It is intuitive to expect that congestion probability of a road should increase 
with the number of congested neighbors, 
Specifically, one would expect that $a_{i,0} \leq a_{i,1} \leq \dots \leq  a_{i,|\pi_i|}$.
However,  the learning strategy of \cite{hofleitner12} doesn't ensure this total ordering.  
Hence, it may be difficult
to interpret real world dependency among neighboring roads from learnt parameters. 
\end{itemize}

{\bf Proposed Transition CPD:}
To alleviate the above short-comings, we propose to use a NoisyOR CPD \cite{Koller09} for modeling
state transitions. If $Y\in\{0, 1\}, $ is the output and 
$\underline{X} = (X_1,X_2,\dots X_n), \:\: X_k \in \{0, 1\},$ is the input, 
then the NoisyOR CPD is parameterized by $n+1$ parameters, 
{\it viz.}  $(q_{0},q_{1},\dots q_{n}), 0\leq q_i \leq 1$, referred to as inhibitor probabilities.
The CPD is given by:
\begin{equation}
\vspace{-0.030in}
 P(Y=0|\underline{X}) = q_0\prod_{k=0}^{n}q_k^{X_k}, \:\:\: X_k \in \{0,1\}.
\label{eq:NoisyORCPD}
\end{equation}
When $q_0 = 1$ and $q_k = 0, \:\: \forall k>0$, we have the noiseless OR function. When one or more of 
the $q_k$s are non-zero, this CPD allows for a non-zero chance of the output becoming $0$ in spite of one 
or more high inputs.   {\em In our context, with $q_0$ clamped to $1$,  each $q_k$ exactly captures the  
chance of output being $0$  (and hence the chance of congestion) when only the
$k^{th}$ neighbor is congested}. Hence, the NoisyOR (unlike SatPat) captures influence of neighboring 
links  in an  independent  and link-dependent fashion --  with $q_k$ representing
the extent of influence from the $k^{th}$ neighbor.   As the number of congested inputs (neighbors) 
go up, the chance of un-saturation goes down as  is evident from eq.~\ref{eq:NoisyORCPD} . 
Hence it also {\em captures the intuition of congestion probability increasing with the number of
congested neighbors in the previous time step}. 
$(1-q_0)$  captures the chance of congestion getting 
triggered spontaneously at a link  (while all its neighbors are uncongested).

{\bf Alternative representation for NoisyOR:} 
The NoisyOR comes from the class of ICI (Independence of Causal Influence) models \cite{Heckerman94} 
and  can be viewed as in Fig.~\ref{fig:NoisyOR} . On each
input line $X_k$, there is a stochastic line failure function, whose output is $Z_k$. The deterministic OR 
acts on the $Z_k$s. 
When the input $X_k$ is zero, the line output $Z_k$ is also zero. When $X_k=1$,  with inhibitor 
probability $q_k$, line failure happens -- in other words, $Z_k$ is zero. The bias term $q_0$ 
controls the chance of the output being $1$ in spite of all inputs being off. 
It is easy to check that  CPD in 
Fig.~\ref{fig:NoisyOR} is  given by eq.~\ref{eq:NoisyORCPD} .   
\newcomment{
\begin{eqnarray} 
&&P(Y=0|\underline{X}=(x_1,x_2 \dots x_n)) \\ \nonumber
& = &P(\underline{Z} = \underline{0}|\underline{X}=(x_1,x_2\dots x_n)) \\ \nonumber
& = & P(Z_0 = 0)\prod_{k=1}^{n} P(Z_k = 0|X_k = x_k) \\ \nonumber
& = & q_0\prod_{k=0}^{n}q_k^{X_k} 
\end{eqnarray}
This first equality is due to $Y$ being a deterministic OR of all the components of $\underline{Z}$. The next 
follows from the factorization definition of a belief network.
}

Under the NoisyOR CPD, the term which models the hidden state transitions can be expressed as 
$A({\bm{\eta}}^{i,t-1}, \bar{\bm{\eta}}^{i,t-1}, s^{i,t})$ where,
${\bm{\eta}}^{i,t-1} = \left[{\eta}^{i,t-1}_1,{\eta}^{i,t-1}_2,\dots, {\eta}^{i,t-1}_{|\pi_i|}\right]$
with ${\eta}^{i,t-1}_j$ representing the actual state of $i$'s  neighbor $j$ at time $t-1$. 
Similarly, $\bar{\bm{\eta}}^{i,t-1} =
\left[\bar{\eta}^{i,t-1}_0,\bar{\eta}^{i,t-1}_1,\dots \bar{\eta}^{i,t-1}_{|\pi_i|}\right]$
with $\bar{\eta}^{i,t-1}_j$  denoting the new random variable introduced 
via the representation of Fig.~\ref{fig:NoisyOR}. 
Note that $\bar{\bm{\eta}}^{i,t}$ is of length $|(\pi_i + 1)|$ while that of $\bm{\eta}^{i,t}$ is  $|\pi_i|$. 
Based on  Fig.~\ref{fig:NoisyOR}, we can write $A({\bm{\eta}}^{i,t-1}, \bar{\bm{\eta}}^{i,t-1}, s^{i,t})$, the transition factor, as follows:
\begin{equation}
\begin{split}
& A(\bm{\eta}^{i,t-1},\bm{\bar{\eta}}^{i,t-1},s^{i,t})  \\
= & \,P(\bar{\eta}^{i,t-1}_0)P(s^{i,t}|\bar{\bm{\eta}}^{i,t-1})\prod_{j=1}^{|\pi_i|} P(\bar{\eta}^{i,t-1}_j|\eta^{i,t-1}_j) 
\\
= &  \,q_{i,0}^{(1-\bar{\eta}^{i,t-1}_0)} p_{i,0}^{\bar{\eta}^{i,t-1}_0} \mathds{1}_{\left\{\textrm{OR}(\bar{\bm{\eta}}^{i,t-1}) 
= s^{i,t}\right\}}\\ 
& \,\prod_{j=1}^{|\pi_i|} (q_{i,j})^{\eta^{i,t-1}_j(1-\bar{\eta}^{i,t-1}_j) }
(p_{i,j})^{\eta^{i,t-1}_j\bar{\eta}^{i,t-1}_j } 
\label{eq:TransProbFactorNoisyOR}
\end{split}
\end{equation}
\newcomment{
\begin{equation}
\label{eq:TransProbFactorNoisyOR}
\begin{split}
& A(\bm{\eta}^{i,t-1},\bm{\bar{\eta}}^{i,t-1},s^{i,t}) = P(\bar{\eta}^{i,t-1}_0)P(s^{i,t}|\bar{\bm{\eta}}^{i,t-1})\prod_{j=1}^{|\pi_i|} P(\bar{\eta}^{i,t-1}_j|\eta^{i,t-1}_j) \\
&=   q_{i,0}^{(1-\bar{\eta}^{i,t-1}_0)} p_{i,0}^{\bar{\eta}^{i,t-1}_0} \mathds{1}_{\left\{\textrm{OR}(\bar{\bm{\eta}}^{i,t-1}) = s^{i,t}\right\}} 
\prod_{j=1}^{|\pi_i|} (q_{i,j})^{\eta^{i,t-1}_j(1-\bar{\eta}^{i,t-1}_j) } (p_{i,j})^{\eta^{i,t-1}_j\bar{\eta}^{i,t-1}_j }\\ 
\end{split}
\end{equation}
}
where $p_{i,j}= 1-q_{i,j}$ and $q_{i,j}$ is the probability that congestion at time step $t-1$ in the $j^{th}$ neighbor of link $i$ 
does not influence $i$ in time step $t$.
Similar to the SatPat CPD (eq.\ref{eq:TransProbFactorSatPat}), {\it eq.\ref{eq:TransProbFactorNoisyOR}   demonstrates that a typical transition  factor in the DBN under 
the NoisyOR CPD also 
belongs to the exponential family.   This in turn makes
M-step of EM learning feasible in closed form as explained later. }

{\bf Complete data likelihood under NoisyOR:}
If $\B{s}$ denotes the state of all links across all time, and $\B{y}$ denotes the set of all travel time observations 
across all vehicles over time $t = 1, \cdots, T$, the complete data likelihood   
is given by:
\begin{equation}
\vspace{-0.114in}
\begin{split}
 p(\B{s},\B{y}| \bm{\theta}) = & \prod_{\substack{t=2\dots T \\ i\in \mathcal{I} }} 
 A(\bm{\eta}^{i,t-1}, \bar{\bm{\eta}}^{i,t-1}, s^{i,t}) 
  \times  \\ 
  & \prod_{\substack{t=1\dots T \\ k=1\dots N_t^v}} f(y_t^k|s^{L_t(k),t}) \times  \prod_{i\in \scr{I}} c^i(s^{i,1}) 
\end{split}
\label{eqn:datalikelihood}
\end{equation}
\newcomment{
\begin{equation}
\begin{split}
& \prod_{\substack{t=2\dots T ,\,  i\in \mathcal{I} }} 
 A(\bm{\eta}^{i,t-1}, \bar{\bm{\eta}}^{i,t-1}, s^{i,t})   \times  \prod_{\substack{t=1\dots T,\,  k=1\dots N_t^v}} f(y_t^k|s^{L_t(k),t}) \times  \prod_{i\in \scr{I}} c^i(s^{i,1}) 
\end{split}
\label{eqn:datalikelihood}
\end{equation}
}
where $c^i(0)$ is the marginal probability of link
$i$ being uncongested at time $1$. We subsume this into $A(.,.,.)$ by constraining $c^i(0) = q_{i,0}$. This is same as assuming all links
start at time $0$ uncongested. Note that,  here $\bm{\theta} = (\bm{q}, \bm{\mu}, \bm{\sigma})$, where $\bm{q}$ refers to all 
NoisyOR parameters of each of the links.  
\section{Learning}
\label{sec:Learning}
An EM approach  (App.~\ref{sec:EMbasics}) is employed  which  is a standard iterative process involving two steps  at each iteration. The E-step computes expectation of complete data log-likelihood ($Q$-function in short) at the current parameter values, while the $M$-step
updates parameters by maximizing the $Q$-function.
When the complete data log likelihood belongs to the exponential family, then learning gets 
simplified \cite{Bishop06,Koller09}.
\newcomment{
Under completely observable data, the maximum likelihood estimate (MLE) is a function of the data via the  sufficient statistics
(SS) alone
and typically there exists a closed form solution. 
In a partially observable data setting also, the EM algorithm gets simplified.} The $E$-step involves just computing the Expected Sufficient
Statistics (ESS). 
The $M$-step typically consists of evaluating an algebraic expression based on the closed form maximum likelihood estimate (MLE) under completely observable 
data, in which  SS is replaced by ESS.  

{\bf E-step:} E-step which involves  ESS computation, is actually performing inference on a belief network.
 Exact inference in multiply connected belief networks  is known to be NP-hard \cite{Cooper90}.  
 Since our DBN is also multiply connected with a large number of links, exact inference would lead to 
 unreasonable run times. Hence we use a 
 sampling based  approximation algorithm for inference \cite{hofleitner12}.     
Specifically, we use a particle filtering approach.  This involves storing and tracking a set of samples or particles. For each particle $r$, we start off with a vector of 
uncongested initial states for
all the links. At each time step $t$, we  grow each particle (sample) based on the current transition probability parameters(NoisyOR or SatPat). Each particle in state 
$s^{i,t}_r$ is now weighted by $\prod_{ k=1\dots N_t^v}
f(y_t^k|s^{L_t(k),t}_r)$. An additional resampling of the particles based on these weights (normalized) is performed to avoid degeneracy.  
The required ESS (described above) are then estimated from these sample paths. {\em As the name filtering
indicates, the ESS at time $t$ is calculated based on observations up to time $t$, namely $\B{y}^t$, rather than all observations.}
The ESS associated with observation parameters turns out to be $P(s^{L_t(k),t} = \bm{z}|\B{y}^t,\bm{\theta}^{\ell})\;\forall t,k,\bm{z}$.
Here, $\bm{z}$ refers to a binary vector of length $|L_t(k)|$. 
\subsection{M-step Update for DBN model }
\label{sec:Mstep}
{\bf Observation updates:} From eq.~\ref{eqn:datalikelihood} , it follows that $Q$-fn for the DBN model involves a sum of two terms: one exclusively a function of 
observation parameters ($\bm{\mu}, \bm{\sigma}$) and the other only of the transition parameters ($\bm{q}$) for NoisyOR. 
Hence the joint maximization over ($\bm{\mu}, \bm{\sigma}$) and ($\bm{q}$) gets decoupled. 
High time-resolution GPS observations  
are used to learn a 2-component Gaussian mixture at each link, which gives the  means and variances of the individual link travel times.  For convenient optimization, 
the variances thus obtained can be fixed and 
learning performed only over $\bm{\mu}$  as carried out  in \cite{hofleitner12}. However, one still needs iterative optimization owing to the complexity of the term
involved.

\newcomment{
\noindent
{\bf Observation updates:}  
 This term is non-convex in both $\bm{\mu}$
and $\bm{\sigma}$, but convex in $\bm{\mu}$ alone. So at each link, the observations  
are used to learn a 2-component Gaussian mixture which gives the link-associated means and variances.
As done in \cite{hofleitner12}, the variances thus obtained can be fixed and
learning carried only over $\bm{\mu}$ for convenient optimization. 
}

{\bf Proposed transition parameters updates:} 
Maximization of the second term involving hidden state transition parameters
leads to an elegant closed-form estimate of the transition parameters for the proposed NoisyOR transitions.
This is mainly because each  factor belongs  to the exponential family. 

\begin{proposition}
\label{prop:MUpdate}
Given the observations $\B{y}$ and parameter estimate after the $\ell^{th}$ EM-iteration, $\bm{\theta}^{\ell}$, the next set of transition parameters are obtained as follows.
\newcomment{
\begin{eqnarray}
q_{i,j}^{\ell + 1} & \propto & \sum_{t=2}^{T}P(\eta^{i,t-1}_j = 1,\bar{\eta}^{i,t-1}_j = 0|\B{y},\bm{\theta}^{\ell})  \nonumber \\
p_{i,j}^{\ell + 1} & \propto & \sum_{t=2}^{T}P(\eta^{i,t-1}_j = 1,\bar{\eta}^{i,t-1}_j = 1|\B{y},\bm{\theta}^{\ell}) 
\label{eq:NoisyORMUpdate}
\end{eqnarray}
}
\begin{equation}
\begin{split}
q_{i,j}^{\ell + 1} & \propto  \sum_{t=2}^{T}P(\eta^{i,t-1}_j = 1,\bar{\eta}^{i,t-1}_j = 0|\B{y},\bm{\theta}^{\ell})  \\
p_{i,j}^{\ell + 1} & \propto  \sum_{t=2}^{T}P(\eta^{i,t-1}_j = 1,\bar{\eta}^{i,t-1}_j = 1|\B{y},\bm{\theta}^{\ell}) 
\label{eq:NoisyORMUpdate}
\end{split}
\end{equation}
where proportionality constants are same. Similarly for $j=0$, the $M$-step updates are:
 \begin{equation}
\begin{split}
q_{i,0}^{\ell + 1} & \propto  \sum_{t=1}^{T}P(\bar{\eta}^{i,t-1}_0 = 0|\B{y},\bm{\theta}^{\ell}) \\ 
p_{i,0}^{\ell + 1} & \propto  \sum_{t=1}^{T}P(\bar{\eta}^{i,t-1}_0 = 1|\B{y},\bm{\theta}^{\ell}) 
\label{eq:BiasUpdate}
\end{split}
\end{equation}
\end{proposition}
Please refer to App.~\ref{sec:PropProof} for a proof. The proof involves computing the $Q$-function for the proposed NoisyOR CPD and maximizing it in closed form.  
\newcomment{
Recall from end of Section~\ref{sec:Model}, that we start with all links in an uncongested state at time $0$, i.e. $\bm{\eta}^{i,0} = \bm{0}\,\forall i$. 
Hence, from fig.~\ref{fig:NoisyOR}, it is easy to see that $s^{i,1}$ is same as $\bar{\eta}^{i,0}_0$. Hence  $\bar{\eta}^{i,0}_0 $ can be replaced by $s^{i,1}$ in 
eq.\ref{eq:BiasUpdate}.
}  
The above ESS  are actually computed conditioned on $\B{y^t}$ (observations upto time t) and not $\B{y}$, via particle filtering  
as explained in the E-step. Attempting smoothing which is exact,  using all the observations $\B{y}$, would lead to  unreasonable space complexities, given the large
number of links.
The above updates are for data observations from a single day. They can readily be extended to multiple days and handled efficiently in a parallel fashion as explained in
App.~\ref{sec:MultDays} . For a comparison of complexities between NoisyOR and SatPat, refer to App.~\ref{sec:complexity} .


\section{Prediction}
\label{sec:prediction}
\SetKw{KwAnd}{and}
\SetKwFunction{GetPotentialCandidates}{GetPotentialCandidates}
\SetKwFunction{suff}{suff}
\SetKwData{TRUE}{TRUE}
\SetKwData{FALSE}{FALSE}
\SetKwData{CS}{CurSuff}
\SetKwData{CSt}{CurSt}
\SetKwData{fs}{FutStep}
\SetKwData{TT}{MTT}
\begin{algorithm*}[t]
\label{algo:prediction}
\caption{Compute expected travel time of an aribitrary length query route}
\KwIn{$\theta^{\star}$, Query Path $\Gamma = [i_1,i_2,\dots i_{|\Gamma|}]$, $\alpha_s$ - fractional distance of $x_s$ from downstream end of $i_1$.  }
\KwOut {Mean Travel time ($\TT$) of traversing $\Gamma = [i_1,i_2,\dots i_{|\Gamma|}]$, starting at $t\Delta$ from $x_s$ on $i_1$. }
Initialize $\TT=0$, $\CS = \Gamma$, $\CSt = \alpha_s$, $\fs = 1$, $\scr{P}=$ Set of particles grown upto $t$\;
\While{$\CS \neq  \phi$}{
                                Grow all particles in $\scr{P}$ by one step (either as per NoisyOR or SatPat transitions)\;
	$L:=\ell$-length prefix path of \CS, say $[i'_1,i'_2,\dots i'_{\ell}]$.  
$\bm{b}_{k-1}$ := $\ell$-length binary representation of $(k-1)$.\\ 
	$\bm{M}_{\ell}(k)  := \CSt*\mu^{i'_1,\bm{b}_{k-1}(1)} + \sum_{j=2}^{\ell} \mu^{i'_j,\bm{b}_{k-1}(j)}$,   
$\bm{p}_{\ell}(k):= P(s^{L,t+\fs} = \bm{b}_{k-1}|\B{y^t},\theta^{\star})$, ($2^{\ell}$-length  vectors).\\ 
                                        \If{$\exists$ an $\ell$ s.t. $\bm{p}_{\ell}^{T}\bm{M}_{\ell}>\Delta$  }{
                                        Compute the least $\ell$ (say $c$)  using binary search  (Use $\scr{P}$, the current set of particles to compute $\bm{p}_{\ell}$) \;
\If{$c>1$}{
$\CSt \leftarrow 1- \{(\Delta - \bm{p}_{c}^T\bm{M}_{c}^{e-})/\bm{p}_c^T\bm{M}_{c}^{e}\}$, where $\bm{M}_{c}^{e-}$, $\bm{M}_{c}^{e}$ are $2^{c}$-length  vectors\;
$\bm{M}_{c}^{e-}(k) := \CSt*\mu^{i'_1,\bm{b}_{k-1}(1)} + \sum_{j=2}^{c-1} \mu^{i_j,\bm{b}_{k-1}(j)}$,
$\bm{M}_c^e = [\mu^{i_c,0}\,\mu^{i_c,1}\,\mu^{i_c,0}\,\mu^{i_c,1}\, \dots \mu^{i_c,0}\,\mu^{i_c,1}]^{T}$. \\
$\CS \leftarrow$ suffix of \CS (from $c$); $\TT \leftarrow \TT + \Delta$; $\fs\leftarrow \fs + 1$\;
}
\lElse{
$\CSt \leftarrow \CSt(1 - (\Delta/\bm{p}_c^T\bm{M}_{c}^{e}))$;   $\TT \leftarrow \TT + \Delta$; $\fs\leftarrow \fs + 1$
}
                                        }    
\lElse{$\TT \leftarrow \TT + \bm{p}_{|\CS|}^{T}\bm{M}_{|\CS|}$; $\CS = \phi$
}
                        }    
                        \Return{$\TT$}
\end{algorithm*}
Formally,  given $\bm{\theta}^{\star}$ (learnt DBN  parameters from historical data) and
current  probe vehicle observations  up to time $t\Delta$ (or time bin $t$), the objective is to predict the
travel time of a vehicle that traverses a specified trajectory (or path) $\Gamma = [i_1,i_2,\dots i_{|\Gamma|}]$
starting at say $t\Delta$ (from time bin $(t+1)$).
Existing works \cite{herring10,hofleitner12}  estimate the travel
time along $\Gamma$ under the assumption that it is lesser than $\Delta$ (or one time step).
However, in general, the travel times for a trajectory can be much more than $\Delta$. 

{\bf Challenge:}
As the DBN evolves every $\Delta$ time units, the state of the DBN estimated at $(t+1)^{th}$ time bin
can be used to predict the network travel times only in the associated time interval $[t\Delta, (t+1)\Delta)$. 
If the given trajectory $\Gamma$ is not fully traversed by $(t+1)\Delta$, 
the DBN's state has to be advanced to time epoch $t+2$. The 
estimated network state at $t+2$ 
should now be used to predict the network travel time in the interval $[(t+1)\Delta, (t+2)\Delta)$, and so on.
In other words, the task of predicting the travel time along $\Gamma = [i_1, i_2, \dots i_{|\Gamma|}]$
now gets transformed to the task of partitioning $\Gamma$ to contiguous trip segments 
$u_1, u_2, \dots, u_M, $ such that the expected travel time of segment $u_j, \: 1 \leq j \leq  (M-1)$,  is
$\Delta$ time units as predicted using the hidden network state estimated at time epoch $(t+j)$;
$u_{M}$ corresponds to the final trip segment in $\Gamma$  whose expected travel time is less than or equal to
$\Delta$.

{\bf Approach:} 
Without loss of generality, we 
assume that the end point $x_e$ of $\Gamma$ coincides with the end of link $i_{|\Gamma|}$. 
Algorithm~\ref{algo:prediction}  describes the procedure to predict the mean travel time, \TT, of $\Gamma$.  App.~\ref{sec:CorrProof} describes its correctness  proof.
\CS refers to the currently remaining suffix of $\Gamma$. 
\CSt is the fractional distance of the start 
point of \CS  from the downstream intersection. $\bm{p}_{\ell}^{T}\bm{M}_{\ell}$ - Expected travel time of traversing a $\ell$-length prefix segment, $L = [i'_1,i'_2,\dots
i'_{\ell}]$, of \CS. The idea is to first narrow
down on the earliest $\ell$ (say $c$) at which  $\bm{p}_{\ell}^{T}\bm{M}_{\ell} > \Delta$. Subsequently, we need the exact position on $i'_{c}$ upto which expected travel time
is exactly $\Delta$. The main component of the proof explained in  App.~\ref{sec:CorrProof} involves how to arrive  at this exact position via a closed-form. This is 
utilized in lines $9$
and $13$ of Algorithm~\ref{algo:prediction}. \fs keeps track of additional number of time steps, particles are currently grown upto.  

\section {Experimental Results }
\label{sec:Simulation}
\cite{hofleitner12}, which propose the SatPat CPD clearly demonstrate how their approach outperforms baseline approaches based on time series ideas.
Given this and comments made earlier in Sec.~\ref{sec:RelWork}, we compare our proposed method with SatPat method only. We first test the efficacy of the methods on 
synthetic data. This is to better understand the maximum performance difference that can occur between the two approaches.

\newcomment{
We test the performance of our proposed approach to predict travel time on both synthetic and real world traffic traces.
We compare the performance our approach with the one suggested in \cite{hofleitner12}. 
We do not compare it with the approach of \cite{hofleitner12a} as it needs an exponential (in the number of neighbors) number of 
parameters to be learnt.  Further, we also do not compare with baseline approaches involving time-series models as the approach in \cite{hofleitner12} and
\cite{hofleitner12a} have been shown to outperform such approaches. We have also verified the correctness of our implementation via likelihood checks as explained in the
appendix.
}

 We implement learning
by updating only the $\bm{q}$ (NoisyOR case) or $\bm{a}$ (SatPat case)  parameters. During learning on synthetic data, we 
fix the observation parameters to the true values with which the data was generated. For ease of verification and since our contribution is in the M-update of hidden state
transition parameters, we stick to this here. However, it is straightforward to include $\bm{\mu}$ as well in the iterative process as
described in Sec.~\ref{sec:Mstep}. The real data we consider in this paper   is  high time resolution probe vehicle data, where one can obtain 
independent samples of individual link times and learn a 2-component Gaussian mixture at each link. Another justification of our approach could be that there may
not be a necessity to update $\bm{\mu}$ and $\bm{\sigma}$ once learnt via high time resolution GPS data.  

We briefly summarize the synthetic data generation setup here and point the reader to App.~\ref{sec:SyntheticData} for additional details. {\em The main idea is to use the DBN
model of Sec.~\ref{sec:Model} with NoisyOR transitions and Gaussian travel times to generate trajectories.}  
The generator takes as input a road network's neighborhood structure and individual link lengths. The  DBN structure is fixed from neighborhood information. 
The NoisyOR CPD gives a nice handle  to embed a variety of congestion patterns. We choose CPD parameters  to embed short-lived and long-lived  congestions. 
The chosen synthetic network has $20$ links with  gridded one-way roads mimicing a typical downtown area. We chose $8$ probe vehicles to circularly ply around the
north-south region while another $8$ along the east-west corridor.   
\subsection{Results on synthetic traces}
\begin{figure}
\center
\subfigure[Mean Absolute Error  with increasing trip duration.]{
\includegraphics[width=1.55in,height=1.15in]{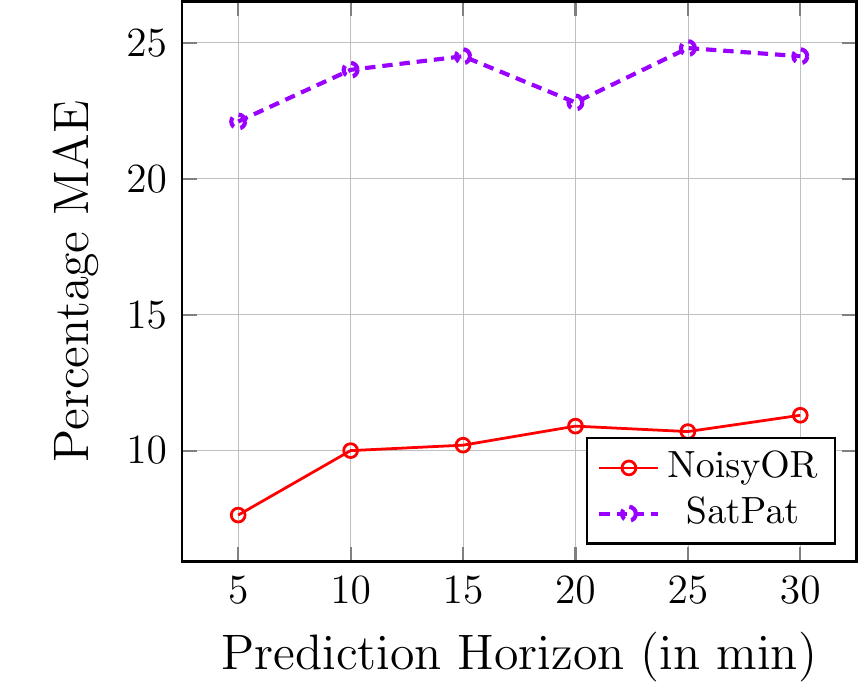}
\label{fig:MAErrorPersisting}
}
\subfigure[Performance  at maximum error difference.]{
\includegraphics[width=1.55in,height=1.15in]{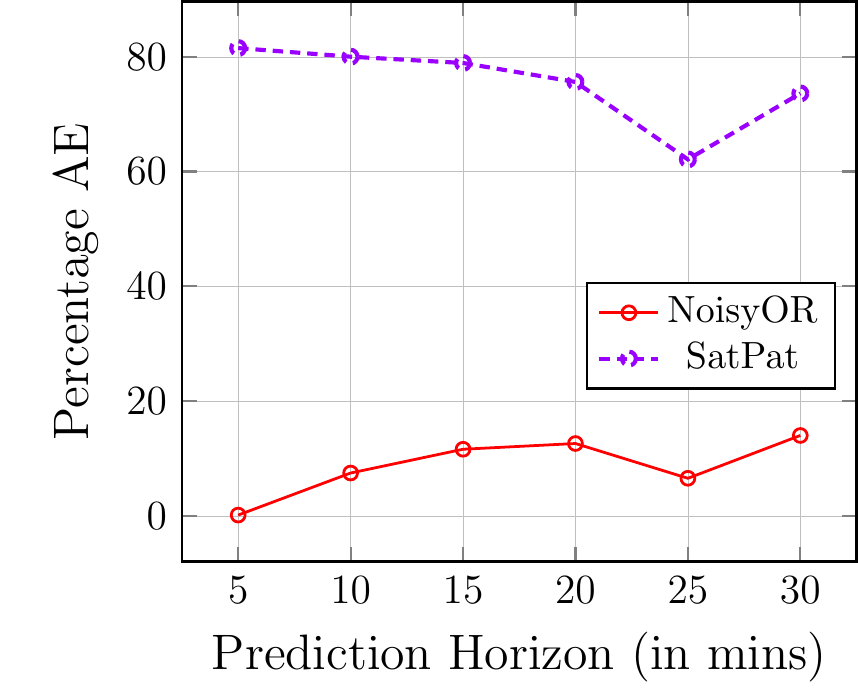}
\label{fig:WCErrorPersisting}
}
\caption{Error vs prediction horizon (True Trip Duration) - long-lived congestions, $\Delta=5$ mins.}
\label{fig:Persisting}
\vspace{-0.12in}
\end{figure}


We compare prediction error between proposed and existing methods as (true) trip duration  is gradually 
increased. Specifically, we use the clearly distinct NoisyOR  learning scheme (proposed) and SatPat learning  scheme (existing) for comparison
(Sec.~\ref{sec:Learning}). 
{\em For prediction however, we emphasize that the algorithm used for comparisons here (for both NoisyOR and SatPat schemes) is not an existing algorithm but
rather a generic 
one proposed here in 
Sec.~\ref{sec:prediction} which can tackle trips of arbitrary duration.} 
We randomly pick from the testing trajectories of each of the $16$ probe vehicles, distinct  non-overlapping trips of a fixed duration. 
We provide results of persisting (OR long-lived) congestion alone here. 
Results on short-lived congestion were found to be similar. 

Each point in fig.~\ref{fig:MAErrorPersisting} shows (Relative) Mean Absolute Error (MAE), obtained by averaging across all the distinct randomly chosen trips of a  fixed 
duration (true  trip time). `Relative' here refers to error normalized by the true trip time.
As true trip time (or prediction horizon) of the chosen trajectories is increased, the  (MAE) also increases as intuitively expected. We also find that NoisyOR consistently gives more accurate 
predictions than SatPat justifying the need to model the varying influences of individual neighbors.
For every prediction horizon,
we also look for a trip on which difference in prediction errors between the proposed and existing approach is maximum. Fig.~\ref{fig:WCErrorPersisting} gives the performance of both NoisyOR and
SatPat with the maximum difference in prediction error, for a given prediction horizon.  
We see that prediction error difference  can be as high as 70\%, with NoisyOR being  
more accurate. 
\newcomment{
{\bf Long-lived congestion:} Under persisting congestion, from Fig.~\ref{fig:MAErrorPersisting}, we find that for increasing prediction intervals, the prediction error does not increase as much as in the short-lived congestion scenario for both the 
approaches.  As before, we find that NoisyOR consistently outperforms SatPat and the prediction error in SatPat is 
more than twice that of NoisyOR.
The results in Fig.~\ref{fig:WCError} indicate that in the worst case, the difference in prediction error can be
 $70\%$ or more  in favor of the proposed NoisyOR method. 
}
{\em Overall, it can be summarized that NoisyOR method's predictions are significantly more accurate than the existing SatPat method.} Further, NoisyOR learnt parameters 
can be interpreted better in real world than SatPat parameters.

\subsection{Results on real-world probe vehicle data}
\begin{figure}[]
\center
\subfigure[Empirical CDF]{
\includegraphics[width=1.55in,height=1.15in]{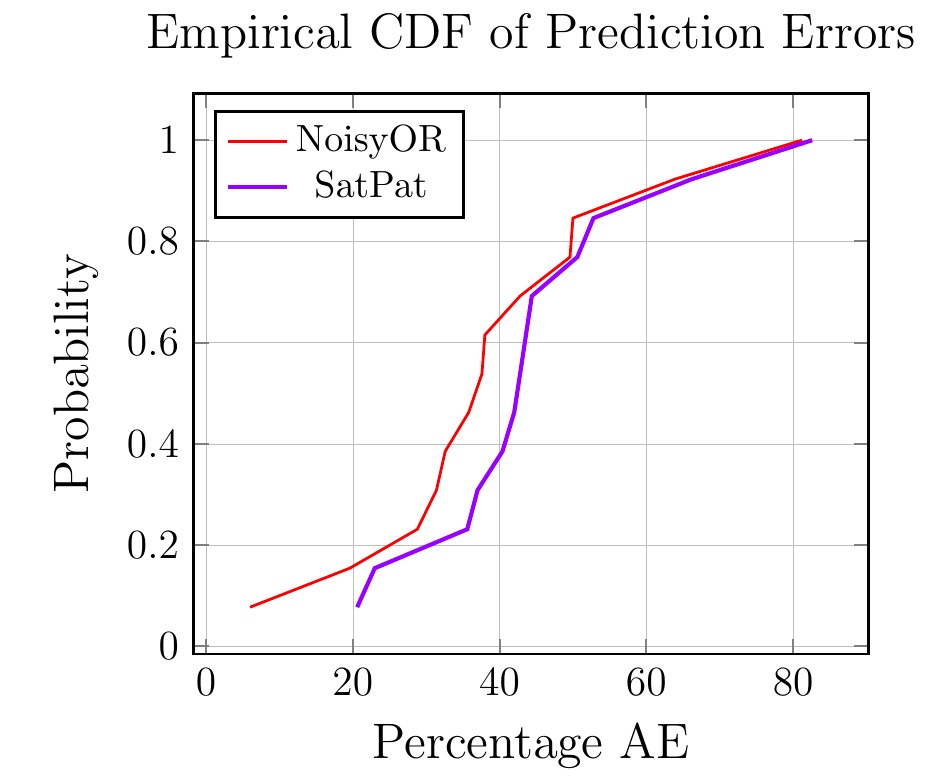}
\label{fig:CDFPorto}
}
\subfigure[Error  time-series]{
\includegraphics[width=1.55in,height=1.15in]{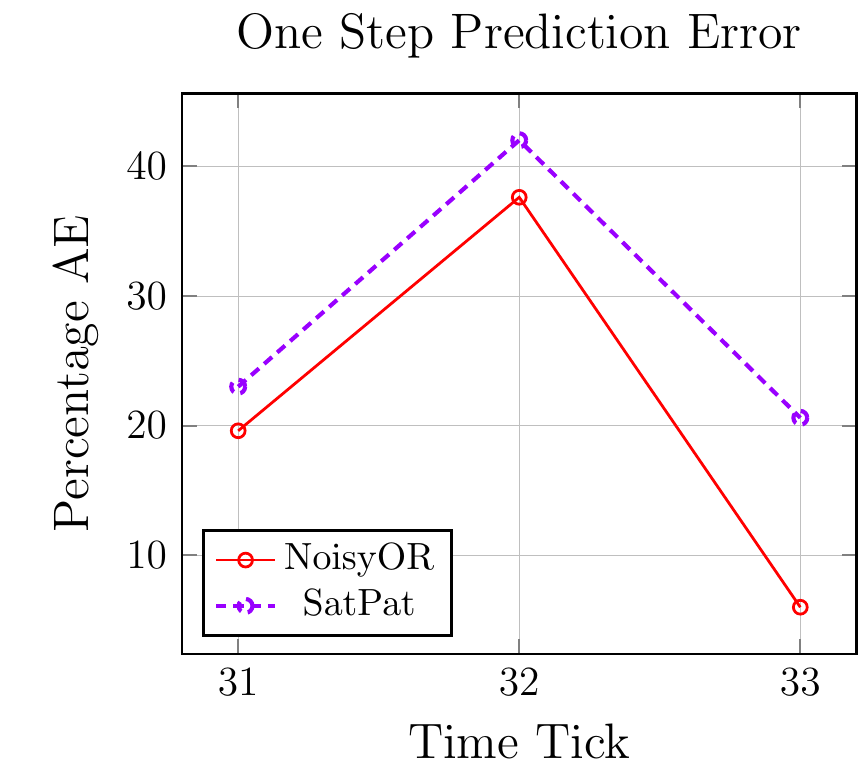}
\label{fig:SampleErrorsPorto}
}
\caption{City of Porto: Test trajectory duration=$\Delta$($5$ min).}
\label{fig:RealDataPorto}
\vspace{-0.15in}
\end{figure}
\begin{figure}[]
\center
\subfigure[Empirical CDF]{
\includegraphics[width=1.55in,height=1.15in]{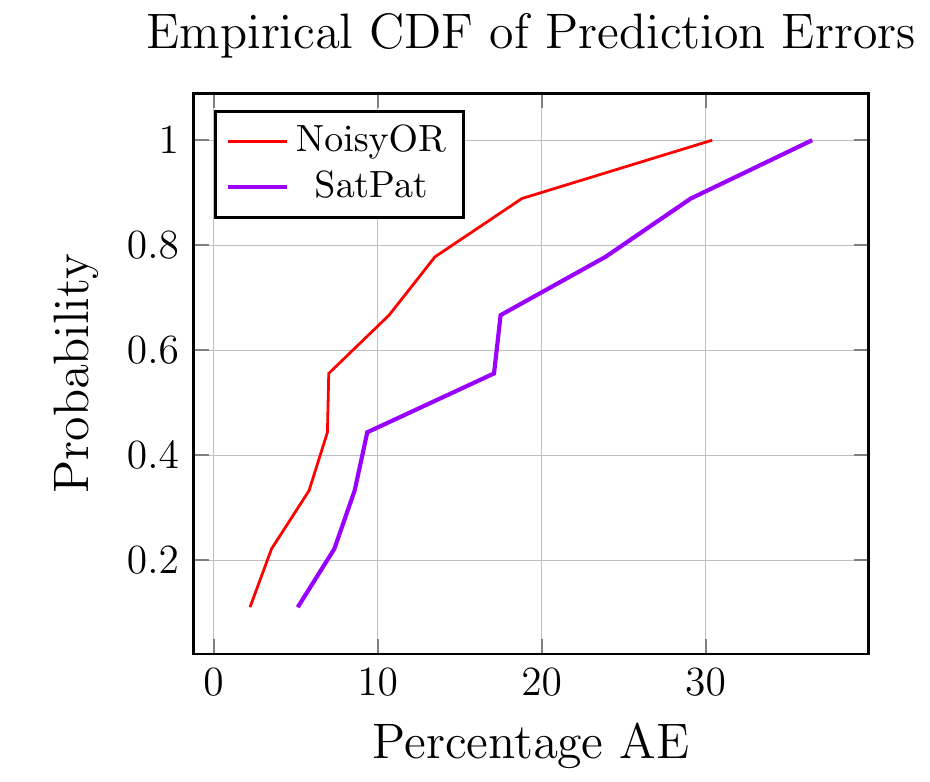}
\label{fig:CDFSFO}
}
\subfigure[Sample of Prediction Errors ]{
\includegraphics[width=1.55in,height=1.15in]{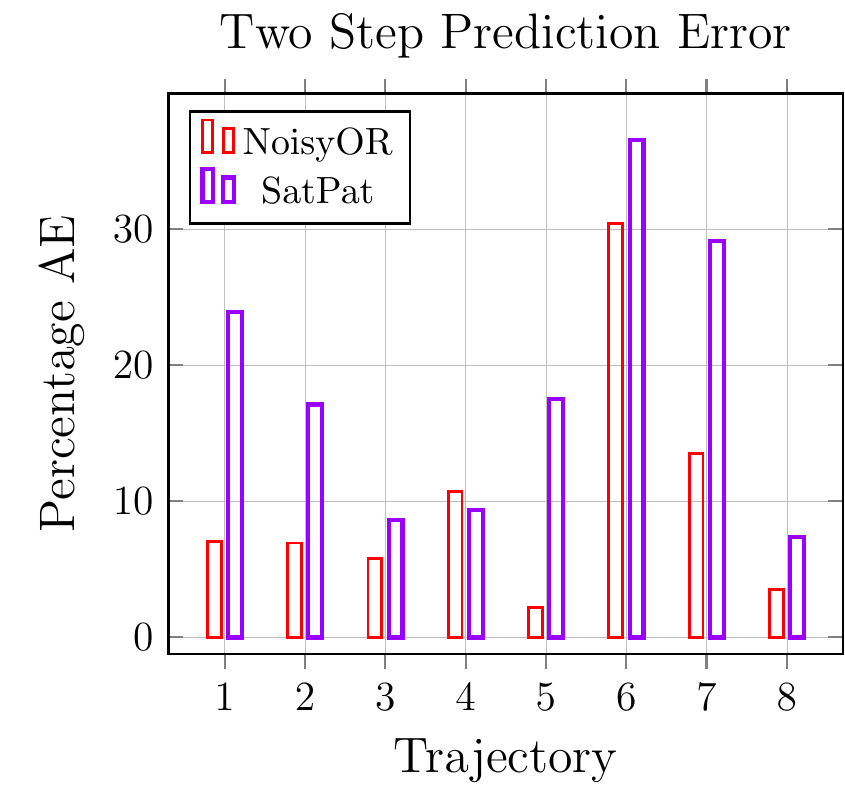}
\label{fig:SampleErrorsSFO}
}
\caption{San Francisco: Test trajectory duration = $2\Delta$($10$ min).}
\label{fig:RealDataSFO}
\vspace{-0.15in}
\end{figure}

{\bf PORTO:} To validate  on real probe vehicle traces,  we first used GPS logs of cabs operating in the city of Porto, Portugal. 
The data was originally released for the ECML/PKDD data challenge  $2015$.    
Each trip entry consists of the start and end time, cab ID and a sequence of GPS
co-ordinates sampled every $15$ seconds.
The GPS co-ordinates in the data are noisy as many of them  map to a point outside the road network. 
The GPS noise was removed using  heuristics such as mapping a noisy point to one or more nearest links on the road network.
The observation parameters $\mu^{i,s}$ and $\sigma^{i,s}$ are learnt for each link $i$ using the high  resolution ($15$ sec) measurements as performed in
\cite{hofleitner12}.
We fix observation parameters  to these values and learn only the transition parameters. 

We choose a connected region of the Porto map which was relatively abundant in car trajectories. 
This region consisted of roughly 100 links. 
App.~\ref{sec:Porto} shows the actual region we narrowed
down to.
We chose a few second order neighbors (neighbor's neighbor) too to 
better capture congestion  propagation. 
{\em Its very likely that a congestion originating at an upstream neighbor of a short link might actually propagate up to a down
stream neighbor of the short link in question within $\Delta$ minutes}. To account for this possibility, we add such second order neighbors (both upstream and downstream)
to the list of original neighbors. 
We quantified short by links $< 75$m in length and pick $\Delta=5$ min.

\newcomment{
We chose a few second order neighbors (in the equivalent graph sense) too to 
better capture the congestion  propagation across the
neighbors. We observed from the link length distribution that there were some links with very low lengths. Given that we are looking at congestion
propagation with DBN time steps of $\Delta = 5$ minutes, {\em its very likely that a congestion originating at an upstream link of a short link might actually propagate up to a down
stream link of the short link in question within $\Delta$ minutes}. To account for this possibility, we also add such second order neighbors (both upstream and downstream)
to the list of original neighbors. For every link $i$ ($< 75$ meters), every down stream neighbor $j$ of $i$ is marked as an additional  neighbor of every upstream neighbor $k$ 
of $i$ and vice
versa.        
}

Trajectories from $4$ p.m. to $9$ p.m. were considered. One can expect the traffic conditions to be fairly stationary in this duration. The traffic patterns
during a Friday evening can be very different from the other weekdays, which is why we treated Fridays separately.  
For sake of brevity, we discuss results obtained on Fridays alone. We trained on the best  (in terms of the number of trajectories) $24$ Fridays.
Training was carried out using both the proposed NoisyOR and existing SatPat CPDs. We tested the learnt parameters on two Fridays.

Fig.~\ref{fig:RealDataPorto} shows the  performance of both the proposed and SatPat method on  trajectories (with true trip time equal to $\Delta$ minutes) 
one time epoch ahead of the current set of observations.   Given the sparse nature of the data obtained, 
we focussed on testing trips of
one $\Delta$ duration. Fig.~\ref{fig:CDFPorto} shows the empirical CDF of the absolute prediction errors (in $\%$). The empirical CDF essentially gives
an estimate of the range of errors both the methods experienced. A relative left shift of the NoisyOR CPD prediction errors indicate a relatively
better performance compared to SatPat. We also observe from the errors that the NoisyOR method has a relative absolute error  of
about $5\%$ lower than SatPat on an average and a relative absolute error of about $14.5\%$ lower in the worst case. Figure~\ref{fig:SampleErrorsPorto} gives a
sequence of (one-step) prediction errors for both methods across a few consecutive time ticks around which data was relatively dense to report
meaningful predictions.  Note that the worst case error of $14.5\%$ was obtained at the $33^{rd}$ 
time tick around which NoisyOR method continues to do better than SatPat.
\newcomment{
We plot the errors over 3-4 consecutive time ticks. Each time-tick $t$ in the figure refers to  a time-epoch up to 
which trajectories are observed and the prediction is carried out on a trajectory one time epoch ahead.  
We observe that the NoisyOR method has a relative absolute
error  of
about $5\%$ lower than SatPat on an average and a relative absolute error of about $13\%$ lower in the worst case.
}

{\bf SAN FRANCISCO:} We also considered  a similar taxi data from a region (please refer to App.~\ref{sec:Porto} for a map view) of the bay area of San Fransico. 
Specifically, we considered  trajectories of $2\Delta$ duration for testing from 
this data. We trained both the NoisyOR and SatPat models on about $11$ days of data collected from this region  of about $275$ links in the evening.  We present results in Fig.~\ref{fig:RealDataSFO} for test trajectories  of
$2\Delta$ duration. As before, the empirical CDF given in Fig.~\ref{fig:CDFSFO} has a relative left-shift in the NoisyOR's CDF, indicative of its
better performance. Further, Fig.~\ref{fig:SampleErrorsSFO} gives the trajectorywise prediction error comparison and an improvement of upto $16.8\%$
was observed in the  worst case and about $6\%$ on an average.

This vindicates that the proposed technique of modeling influences of different roads in propagating traffic congestion can indeed be helpful. 
We also note that the worst case performance different between NoisyOR and SatPat is not as pronounced as in the synthetic traces. This could be
attributed to the one of the following reasons: (i)the underlying congestion propagation characteristics may not be too much
link dependent; (ii) even if the congestion propagation is link dependent, enough samples from probe vehicles may not be present in the available
data logs.


\section{Discussions and Conclusions}
\label{sec:Conclusions}
{\bf NoisyOR Based DBNs in Bioinformatics:} We motivate one other concrete application where our NosiyOR based 
DBNs can be useful. 
Inferring gene regulation networks \cite{Guy} from gene expression data is a very important problem in bioinformatics. Discovering the hidden excitatory/inhibitory interactions amongst the interacting
genes is of interest here. DBN based approaches based on continuous hidden variables have been explored for this problem 
(\cite{Perrin}). The
NoisyOR based DBN and the associated learning algorithm introduced in this paper can be a viable alternative  to infer the underlying gene 
interactions by employing a
fully connected structure among the interacting genes. The learnt $q_{i,j}$ values can potentially indicate the strength of influence. 
We intend exploring this further in our future work.

To conclude the paper, we proposed a balanced data driven approach to 
address the problem of travel time prediction in arterial roads
using data from probe vehicles.
We used a NoisyOR CPD in conjunction with a DBN 
to model the varying degrees of influence a given road  
may experience from its neighbors.
We also proposed an efficient algorithm to learn model parameters. We also proposed an algorithm for predicting travel times of trips of arbitrary duration.
Using synthetic data traces, we quantify the accuracy of the proposed method to 
predict the travel times of arbitrary duration trips under various traffic conditions. 
With the proposed approach, the prediction error
 reduces by as much as $50-70\%$ under certain conditions.
We also tested the performance  on traces of real data and found that the  proposed approach fared better than the existing approaches.
A possible future direction  is to generalize the proposed approach to model road
conditions using more than two states.


\bibliography{Transportation}
\bibliographystyle{aaai}
\newpage
\newpage
\appendix
\section{Table of Symbols}
\label{sec:table}
\begin{table}[h]
\caption{Notation used in this paper.}
\label{tab:notation}
\begin{center}
\begin{tabular}{ll }  \hline
{\bf Symbol} & {\bf Description}   \\  
$\mathcal{I}$ & Set of all links in the road network.\\ 
 $\pi_i$ & Set of links adjacent to road $i$, including itself.  \\ 
 $\Delta$ & Size of a time bin or time between successive  \\ 
& GPS measurements. \\
 $s^{i, t}$ & Random variable representing congestion state  \\ 
& of road $i$ at time step $t$;  $s^{i,t}\in \{0, 1 \}$. \\
 $\mu^{i, s},  \, \sigma^{i, s}$ & Mean and std. deviation of the normally   \\ 
&distributed travel time of link $i$ at state $s$. \\
 $N^v_t$ & Number of active vehicles at time step $t$.  \\ 
  $y_{t}^k$ &  travel time measurement at time step $t$ from the   \\ 
& $k^{th}$ active vehicle; $k\in \{1,\dots, N_t^v\}$. \\
 $L_t(k)$ & Set of links traversed by the $k^{th}$ vehicle   \\ 
& at time step $t$. \\
 ${x}_{s,t}^k$,  ${x}_{e,t}^k$  & Start and end locations   of the $k^{th}$ active vehicle   \\ 
& at time step $t$. \\
 $\tau^{i, t}$ & Actual travel time along link $i$ at time step $t$.  \\ 
$A(.,.)$ & Conditional distribution governing the link state  \\
& transitions in the DBN. \\
 $\bm{\eta}^{i, t}$ & Vector of actual congestion states of   \\ 
& $i$'s neighbors at $t$. \\
 $\bar{\bm{\eta}}^{i, t}$ & Vector representing influence exerted by $i$'s \\
& neighbors on $i$ at time $t$ under NoisyOR CPD. \\ 
 $\bm{\theta}$ & Complete  parameter set  governing the DBN. \\ 
 $\bm{\theta}^{\star}$ & Learnt parameters.  \\ 
$q_{i,j}$  & probability that congestion in the $j^{th}$ neighbor of \\
& $i$ at time step $t-1$ does not 
 influence $i$ at time $t$. \\ 
$a_{i,j}$ &  congestion probability at the  $i^{th}$ link given $j$ of its\\
&  neighbors are congested at the previous  instant. \\ \hline
\end{tabular}
\end{center}
\end{table}

\section{Summary of Travel Time Estimation Methods }
\label{sec:TTE}
Among the travel time  estimation  methods,  
a simple approach  which uses a weighted average of real-time 
and historical data is proposed in \cite{Wenjing09}. A more involved model which exploits the travel time correlation of nearby links to 
again
perform travel time estimation using both historical and real-time travel data is proposed in \cite{El10}. 
The work in \cite{Jenelius13} models link travel
times by breaking them down to segments, assuming (dependent) gaussian travel times on each of these segments. Further a 
complex regression
model  which uses spatial correlation is used for network wide travel time estimation. 	 
A Markov chain approach for arterial travel time
estimation was proposed recently in \cite{Ramezani12}. 
An interesting approach based on tensor decomposition for city-wide travel time estimation based on GPS data has been proposed in \cite{Wang14}. 
The current paper however deals with travel time {\em prediction} using probe vehicle data.

\section{EM algorithm}
\label{sec:EMbasics}
The EM algorithm is a popular method for learning  on probabilistic
models in the presence of hidden variables.The EM algorithm is fundamentally maximizing the observed data likelihood, namely $p(\B{y}|\bm{\theta})$.  Towards this, 
it employs an iterative process involving two steps (E-step 
and M-step) at each iteration.

Given the parameter estimate after the $\ell^{th}$ iteration, $\bm{\theta}^{\ell}$, the $E$-step computes the expected complete data loglikelihood defined as 
follows:
\begin{equation}
Q(\bm{\theta},\bm{\theta}^{\ell}) = \sum_{s}p(\B{s}| \B{y},\bm{\theta}^{\ell})ln(p(\B{y},\B{s}| \bm{\theta}))
\label{eq:Qfn}
\end{equation}
where the expectation
is with respect to the conditional distribution $p(\B{s}|\B{y},\bm{\theta}^{\ell})$ computed at the current parameter estimate.

The $M$-step maximizes the above expected likelihood to obtain the next set of parameters, $\bm{\theta}^{\ell+1}$.  
\begin{equation}
\bm{\theta} ^ {\ell+1} = \arg\!\max_{\bm{\theta}} Q(\bm{\theta}, \bm{\theta}^{\ell})
\end{equation}
The EM-algorithm increases the loglikelihood at each iteration and is guaranteed to converge to a local maximum of the data likelihood.

\section{Proof of Proposition~\ref{prop:MUpdate}}
\label{sec:PropProof}
\begin{proof}
NoisyOR CPD factor $A(.,.,. )$ appearing in eq.~\ref{eqn:datalikelihood} can be expressed in terms of the
associated bernoulli random variables and unknown link transition parameters $(\bm{q}^i,\bm{p}^i)$ as given in  eq.~\ref{eq:TransProbFactorNoisyOR}. 
Taking log and expectation with respect to $p(\B{s}|\B{y},\bm{\theta}^{\ell})$ on both sides of eq.~\ref{eq:TransProbFactorNoisyOR}, we get 
the contribution of this factor in the Q-function (eq.~\ref{eq:Qfn} in App.~\ref{sec:EMbasics}) as follows.
\begin{equation}
\begin{split}
& log \left( E\left[ A(\bm{\eta}^{i,t-1},\bm{\bar{\eta}}^{i,t-1},s^{i,t}) \right] \right)   \\
& = \sum_{j=1}^{|\pi_i|} E\left[\eta^{i,t-1}_j (1-\bar{\eta}^{i,t-1}_j)\right]log(q_{i,j}) 
 + E\left[(\bar{\eta}^{i,t-1}_0)\right]log(p_{i,0})  \\
& +E\left[(1-\bar{\eta}^{i,t-1}_0)\right]log(q_{i,0}) +  \sum_{j=1}^{|\pi_i|} E\left[\eta^{i,t-1}_j \bar{\eta}^{i,t-1}_j\right]log(p_{i,j}) \nonumber
\end{split}
\end{equation}
\newcomment{
\begin{multline}
 log \left( E\left[ A(\bm{\eta}^{i,t-1},\bm{\bar{\eta}}^{i,t-1},s^{i,t}) \right] \right)   
 = \sum_{j=1}^{|\pi_i|} E\left[\eta^{i,t-1}_j (1-\bar{\eta}^{i,t-1}_j)\right]log(q_{i,j}) 
 + E\left[(\bar{\eta}^{i,t-1}_0)\right]log(p_{i,0})  
 +E\left[(1-\bar{\eta}^{i,t-1}_0)\right]log(q_{i,0}) +  \sum_{j=1}^{|\pi_i|} E\left[\eta^{i,t-1}_j \bar{\eta}^{i,t-1}_j\right]log(p_{i,j}) \nonumber
\end{multline}
}
The various expectations can be simplified as:
\newcomment{
\begin{equation}
\begin{split}
\label{eq:simplifications}
E\left[\eta^{i,t-1}_j\bar{\eta}^{i,t-1}_j \right] & =  P(\eta^{i,t-1}_j = 1,\bar{\eta}^{i,t-1}_j = 1|\B{y},\bm{\theta}^{\ell})\\
E\left[\eta^{i,t-1}_j(1-\bar{\eta}^{i,t-1}_j)\right] & =  P(\eta^{i,t-1}_j = 1,\bar{\eta}^{i,t-1}_j = 0|\B{y},\bm{\theta}^{\ell}) \\
E\left[(1-\bar{\eta}^{i,t-1}_0)\right] & = P(\bar{\eta}^{i,t-1}_0 = 0|\B{y},\bm{\theta}^{\ell}) \\
E\left[(\bar{\eta}^{i,t-1}_0)\right]  & = P(\bar{\eta}^{i,t-1}_0 = 0|\B{y},\bm{\theta}^{\ell}) 
\end{split}
\vspace{-0.11in}
\end{equation}
}
\begin{eqnarray}
\label{eq:simplifications}
E\left[\eta^{i,t-1}_j\bar{\eta}^{i,t-1}_j \right] & = & P(\eta^{i,t-1}_j = 1,\bar{\eta}^{i,t-1}_j = 1|\B{y},\bm{\theta}^{\ell}) \nonumber \\ 
E\left[\eta^{i,t-1}_j(1-\bar{\eta}^{i,t-1}_j)\right] & = & P(\eta^{i,t-1}_j = 1,\bar{\eta}^{i,t-1}_j = 0|\B{y},\bm{\theta}^{\ell}) \nonumber \\
E\left[(1-\bar{\eta}^{i,t-1}_0)\right] & = &P(\bar{\eta}^{i,t-1}_0 = 0|\B{y},\bm{\theta}^{\ell})  \nonumber \\
E\left[(\bar{\eta}^{i,t-1}_0)\right]  & = & P(\bar{\eta}^{i,t-1}_0 = 0|\B{y},\bm{\theta}^{\ell}) 
\vspace{-0.11in}
\end{eqnarray}
For a fixed $i$ and $j$ ($j>0$), combining all terms involving $p_{i,j}$ and $q_{i,j}$ from the $Q$-fn, we obtain the following term.
\begin{equation}
 \sum_{t=2}^{T}E\left[\eta^{i,t-1}_j (1-\bar{\eta}^{i,t-1}_j)\right] log(q_{i,j}) + E\left[\eta^{i,t-1}_j\bar{\eta}^{i,t-1}_j \right]log(p_{i,j})
\label{eq:exp}
\end{equation}
where, $c_{i,j} =\sum_{t=2}^{T}E\left[\eta^{i,t-1}_j (1-\bar{\eta}^{i,t-1}_j)\right] $, $d_{i,j} = \sum_{t=2}^{T}E\left[\eta^{i,t-1}_j\bar{\eta}^{i,t-1}_j \right] $ are the ESS associated with the
transition parameters for the NoisyOR CPD.

To maximize the above expression (eq.~\ref{eq:exp}) with respect to $q_{i,j}$ and $p_{i,j}$ with the constraint of $q_{i,j}+p_{i,j}=1$,  the method of Lagrange multipliers 
yields the following Lagrangian 
\begin{equation}
\mathscr{L}( q_{i,j}, p_{i,j},\lambda ) = c_{i,j} log(q_{i,j}) + d_{i,j} log(p_{i,j}) + \lambda (1 - (q_{i,j}+p_{i,j}))
\label{eq:LagrangianNoisyOR}
\end{equation}
Differentiating the above equation and equating the first derivatives to zero, along with eq.~\ref{eq:simplifications}  yields the (unnormalized) closed form solution as given in
eq.~\ref{eq:NoisyORMUpdate}.
Similarly for $j=0$, a similar algebra yields eq.~\ref{eq:BiasUpdate} . 
\end{proof}

\section{Handling data from multiple days:}
\label{sec:MultDays}
The updates in eq.~\ref{eq:NoisyORMUpdate} are for data observations from a single day. In general, we would learn from observations from multiple days 
under an i.i.d assumption between different days. In such a case, the right hand sides of the proportionality equations (eq.~\ref{eq:NoisyORMUpdate}) will be involved in a double summation with the outer summation 
running over the day index similar to HMMs \cite{Bishop06}. The individual term calculation and inner summations can all be carried out in parallel i.e. for each day's data before performing the outer summation across days to compute the 
next step transition parameters. Essentially, the $E$-step which computes ESS by using particle filtering can be carried out in parallel for each day.  Hence the algorithm 
is amenable for parallelization and we  exploit this aspect in our implementation.

\section{Complexity: SatPat vs NoisyOR }
\label{sec:complexity}
In comparison to eq.~\ref{eq:NoisyORMUpdate}, the updates of the SatPat parameters \cite{hofleitner12} in the M-step would be as follows.
\begin{equation}
\begin{split}
a_{i,j}^{\ell + 1}  \propto  \sum_{t=1}^{T}P(\sum_{k=1}^{|\pi_i|}\eta^{i,t-1}_k = j,s^{i,t} = 1|\B{y},\theta^{\ell}) \\ 
b_{i,j}^{\ell + 1}  \propto  \sum_{t=1}^{T}P(\sum_{k=1}^{|\pi_i|}\eta^{i,t-1}_k = j,s^{i,t} = 0|\B{y},\theta^{\ell}) 
\end{split}
\end{equation}
For $j>0$, the first term in the above summation will be zero always as we start with all links in an uncongested state at time $0$.

While growing the particles there is a forward sampling step based on the transition probability structure. The introduction of the additional bernoulli random variables 
$\bar{\bm{\eta}}^{i,t-1}$ in the NoisyOR case at each link
$i$, we
would need to potentially sample once (a random number in $[0,1]$) for each of its neighbors. This means we would need to sample $(|\pi_i| + 1)$ times
in the worst
case. For the same reason, one would also need to store an additional $|\pi_i| + 1$ binary
elements per link and per particle during NoisyOR learning.
On the other hand for the SatPat CPD, one needs to compute the number of saturated neighbors (which is still an $\mathcal{O}(|\pi_i|)$ computation) but sample exactly once (irrespective of the
no. of neighbors) to compute the congestion
state of the link at the next time epoch. Since this is done for every particle, particle filtering run times are slightly higher for the NoisyOR CPD. Also while computing the
ESS, in the E-step of the NoisyOR case, for each particle and each time bin (consecutive time bins), there is a contribution to potentially $|\pi_i|$ no. of ESS
(in the worst case) associated with link $i$. As opposed to this, for the SatPat case, there is a contribution to exactly one ESS as is evident from the above equation.

\section{Correctness proof of Algorithm~\ref{algo:prediction}}
\label{sec:CorrProof}
Given a path $\Gamma = [i_1,i_2,\dots i_{|\Gamma|}]$,    the conditional travel time distribution along $L$ at time $(t+1)$, given the current real time 
observations $\B{y^t}$, is a mixture of Gaussians. The number of components in the mixture would be $2^{|\Gamma|}$, with each component weight 
being $P(s^{L,t+1} = \bm{b}|\B{y^t},\theta^{\star})$, where $\bm{b}$ is a binary string of length $|\Gamma|$ that encodes the states of the individual links. 
Each associated component distribution of the mixture 
will itself be a sum of independent Gaussians with a mean of $\sum_{j=1}^{|\Gamma|} \mu^{i_j,\bm{b}(j)}$. This follows from the conditional independence of the Gaussian travel 
times of links given the underlying state information (Section~\ref{sec:keydiff}).

Consider a path $\Gamma = [i_1,i_2,\dots i_{|\Gamma|}]$ with a starting point $x_s$ on link $i_1$, with $\alpha_s$ the fractional distance of $x_s$ from the downstream
intersection of $i_1$. 
For any prefix subpath of $\Gamma$  of length $\ell$ ($L = [i_1,i_2,\dots i_{\ell}]$, $\ell \leq |\Gamma|$),  we introduce two $2^{\ell}$-length  vectors
$\bm{M}_{\ell}$ and $\bm{p}_{\ell}$. $\bm{M}_{\ell}$ denotes the mean vector of all the components 
of the Gaussian mixture distribution modeling the travel time across $L$. Specifically,  
 $\bm{M}_{\ell}(k)  = \alpha_s\mu^{i_1,\bm{b}_{k-1}(1)} + \sum_{j=2}^{\ell} \mu^{i_j,\bm{b}_{k-1}(j)}$, 
where $\bm{b}_k$ is the $\ell$-length binary representation of 
integer $k$. We further assume $\bm{b}_k(1)$ is the MSB, while $\bm{b}_k(\ell)$ is the LSB of this binary representation. Similarly, $\bm{p}_{\ell}$ is
the vector of mixture weights of this gaussian mixture, where  $\bm{p}_{\ell}(k)
= P(s^{L,t+1} = \bm{b}_{k-1}|\B{y^t},\theta^{\star})$. 

Reweighted particles already spawned till time $t$ are now grown by  one step to compute each of the components of
$\bm{p}_{\ell}$ (line $3$ of Algorithm~\ref{algo:prediction}). In general, all particles currently at $(t+\fs-1)$ are grown by one step as per the state
transition structure (NoisyOR or SatPat).

The mean travel time across any expanding path $L$ will be $\bm{p}_{\ell}^{T}\bm{M}_{\ell}$. 
We look for the least $\ell$ such that $\bm{p}_{\ell}^{T}\bm{M}_{\ell}>\Delta$ (line $6$ of Algorithm~\ref{algo:prediction}). Let $c$ be the least $\ell$ satisfying this. 
The mean travel time along an expanding prefix path $L$ is an increasing function of $\ell$. Hence to compute $c$, binary search would be more efficient
(line $7$ of Algorithm~\ref{algo:prediction}).  
In other words, 
we find the first link $i_c$ along the trajectory $\Gamma$  where the mean travel time from $x_s$ becomes greater than
$\Delta$.
After finding $i_c$, we find the precise location $x_c$ on the link $i_c$ such
that the mean travel time from $x_s$  to this point equals $\Delta$. 
This point $x_c$ can be determined using a closed form expression as shown below 
(line $9$ or line $13$ of Algorithm~\ref{algo:prediction}).  

Let $\alpha_c$ denote the fractional distance of $x_c$ from the upstream intersection of $i_c$. 
Consider a $2^c$-length vector $\bm{M}_{c}^{\alpha}$ ($0\leq \alpha \leq 1$), where  
\begin{equation}
\begin{split}
\bm{M}_{c}^{\alpha}(k) = \alpha_s\mu^{i_1,\bm{b}_{k-1}(1)} + \sum_{j=2}^{c-1} \mu^{i_j,\bm{b}_{k-1}(j)} + \alpha \mu^{i_c,\bm{b}_{k-1}(c)}, \\ k=1,2,\dots,2^{c}.
\label{eq:Mac}
\end{split}
\end{equation}
In the above equation, note that $\bm{b}_{k-1}$ is a $c$-length binary representation of $(k-1)$.
To obtain  $\alpha_c$, we   solve for $\alpha$ in the equation $\bm{p}_c^{T}\bm{M}_{c}^{\alpha}=\Delta$. 
Towards solving this, we first recognize from eq.(\ref{eq:Mac}) that $\bm{M}_{c}^{\alpha} = (\bm{M}_{c}^{e-} + \alpha\bm{M}_{c}^{e})$,
essentially a sum of two other  $2^c$-length vectors, one independent and the other dependent on $\alpha$. Here, each component of 
$\bm{M}_c^{e-}$ can be written as  (
definition on line $10$
of Algorithm~\ref{algo:prediction} )
\begin{equation}
\bm{M}_{c}^{e-}(k) = \alpha_s\mu^{i_1,\bm{b}_{k-1}(1)} + \sum_{j=2}^{c-1} \mu^{i_j,\bm{b}_{k-1}(j)} 
\end{equation}
Further, by our convention, $\bm{b}_{k-1}(c)$ is the LSB of the binary representation of $(k-1)$. Hence we have 
$\bm{M}_c^e = [\mu^{i_c,0}\,\mu^{i_c,1}\,\mu^{i_c,0}\,\mu^{i_c,1}\, \dots \mu^{i_c,1}]^{T}$, which is equivalent to $\bm{M}_c^e(k) := \mu^{i_c,(k-1) \pmod{2}}$  (
definition on line $10$
of Algorithm~\ref{algo:prediction} ). Now substituting for $ \bm{M}_{c}^{\alpha}$ by $(\bm{M}_{c}^{e-} +
\alpha\bm{M}_{c}^{e})$ and a term rearrangement yields 
\begin{equation}
\begin{split}
\alpha_c &= (\Delta - \bm{p}_{c}^T\bm{M}_{c}^{e-})/\bm{p}_c^T\bm{M}_{c}^{e}
\end{split}
\end{equation}
$\alpha_c$ denotes the fractional distance of $x_c$ from the upstream end. Since \CSt captures the fractional distance from the downstream end, we set \CSt to
$1-\alpha_c$ as per line $9$ of Algorithm~\ref{algo:prediction} ). \CS is now set to the suffix of the previous \CS ($\Gamma$ in the first iteration) that  starts from
the $c^{th}$ position. Mean Travel Time \TT is incremented by $\Delta$ as a segment with mean travel time $\Delta$ is consumed (line $11$). 

On the other hand if $c=1$, $\bm{p}_1^T\bm{M}_{1}^{e}>\Delta$. $\bm{p}_1^T\bm{M}_{1}^{e}$ is the time taken to travel the rest of $i'_1$ OR a fractional distance \CSt.    
Therefore in a time of $\Delta$, the fractional distance travelled  is $(\CSt*\Delta/\bm{p}_c^T\bm{M}_{c}^{e})$ to obtain $x_c$. The fractional distance of $x_c$ from the
downstream end of $i'_1$ would be  $\CSt - (\CSt*\Delta/\bm{p}_c^T\bm{M}_{c}^{e})$ (line $13$ of Algorithm~\ref{algo:prediction} ).

If the current suffix $\CS$ is such that $\nexists$ an $\ell$ satisfying  $\bm{p}_{\ell}^{T}\bm{M}_{\ell} > \Delta$, then this means we have hit the last segment and \TT is
accordingly incremented by the mean travel time of \CS, the last segment (line $15$ of Algorithm~\ref{algo:prediction}). 
\newcomment{
Note that the trajectory from $x_s$ to $x_c$ along $\Gamma$ denotes the first trip segment $u_1$, described in the beginning of Section~\ref{sec:prediction}.
Once $u_1$ has been determined, the entire process is repeated on the remaining trajectory by setting $x_s = x_c$ to 
determine $u_2$.
 All the particles currently at $(t+1)$ are grown by one step as per the state
transition structure (NoisyOR or SatPat) and the process continues. The process terminates on a suffix trajectory of $\Gamma$ on (whole of) which the expected travel time is less
than $\Delta$. 
This last segment $u_M$ consumes a
time less than or equal to $\Delta$. The sum of the times consumed by each segment $u_i$ gives the total mean travel time of the path $\Gamma$.
}

\section{Correctness of implementation}
\label{sec:ImpCorr}

Any EM algorithm necessarily increases the likelihood of the data at every iteration. In order to verify this necessary condition, we
additionally
implemented a routine to calculate the likelihood of the data given a current set of transition and observation parameters. The implementation
is
based on the forward-backward algorithm for inference in HMMs\cite{Bishop06}. In the current case, if all the link states at a fixed
time are
stacked together
as a vector, then the sequence of joint vectors of hidden link states exactly forms a markov chain. We would need the transition probability matrix which can
be
calculated based
on the transition CPD used.  {\em Note the observations at each time epoch are not the link travel times
directly but a certain linear combination of one more link travel times depending on the vehicle trajectory.} In addition to the transition
probability matrix one needs to compute the probability 
of having traversed each of these trajectories in a time of $\Delta$ units, given the underlying state vector. This is readily computable as the travel times given the underlying states are 
independent gaussians,
which means  the distribution of their appropriate sums are also gaussian.  Since the no. of states of this  markov chain grows exponentially
with the
number of links,  the transition 
matrix and likelihood computation do not scale with no. of links. However, for tiny networks one can use this to compute likelihoods
and hence verify the correctness of implementation. 

\newcomment{
As an initial toy example, we chose a simple road network with $3$ links of equal length.
The  links are such that $1$ feeds into $2$, $2$ feeds into $3$ and $3$ feeds
into $1$. Given this, each link is a neighbor of the remaining links. We chose to use two floating vehicles initially positioned at the
beginning
of link $1$ and $2$. We generated data using the NoisyOR CPD for state transitions in the links. The parameters of each of these were chosen to
enable embedding   a simple congestion pattern. The pattern is a downstream propagation of congestion germinating at link $1$, moving next
to link $2$ and finally to link $3$ and subsequently dying down. The uncongested and congested mean travel times were chosen to be $1.5$ and
$2.5$ units respectively for all the links with a uniform small variance.  The vehicular trajectory data was generated with these parameters for
$100$ time epochs each across
4 days. 
}

On running learning algorithms (NoisyOR and SatPat) on a 3-link network to learn the transition parameters alone with fixed means and variances (to the true values), results
were
as expected. We observed the log likelihoods of the parameter iterates to always increase and generally converge to the true parameter loglikelihoods.
\newcomment{
We observed convergence of not only the log likelihoods of the parameter iterates to the true parameter loglikelihoods, but also convergence of the
transition parameters sufficiently close to the true parameters for a specific choice of the NoisyOR parameters. In general EM algorithm only ensures likelihood increase at each iteration and
convergence to a
local maximum of the observed data likelihood. 
We carried out these preliminary verifications on both NoisyOR and SatPat implementations.
}

\section{Synthetic Data}
\label{sec:SyntheticData}
Testing the algorithms on synthetic data has the following advantages:  
(a) since the ground truth information is
available in a synthetic setup, we can precisely validate the goodness of learning. 
(b)  it also gives us a handle to compare performance of the proposed and existing approaches under different congestion patterns.
\subsection{Trace Generation Setup }
The synthetic data generator is fed with a road network containing a certain number of
links along with their lengths and a neighborhood structure. Based on this neighborhood structure, 
we feed the generator a transition probability structure that governs the congestion state transitions of the 
individual links from time $t$ to time $t+1$. 
The conditional travel times for each link $i$ and state $s$,  is assumed to be
normally distributed with appropriate parameters $\mu^{i,s}$ and $\sigma^{i,s}$  as described earlier in 
Sec.~\ref{sec:Model} . $\mu^{i,0}$ and $\mu^{i,1}$  capture the average travel times experienced by commuters during congestion and non-congestion
due to intersections or traffic lights. The $\sigma$ parameter captures the continuum of congestion levels actually possible.
The link states are assumed to make transitions  at a time scale approximately equal to $\Delta$. 
\smallskip

\noindent
{\bf Data Generation:} 
One or more probe vehicles traverse a subset of links in a predetermined order repeatedly. The paths and vehicles are so chosen that there is  sufficient coverage of all
the links across space and time. The link states are stochastically sampled with time
as per the prefixed transition probability structure. Given the state of all the links at a
particular time epoch, we now describe how the trajectories are generated for the probe vehicles.
Let us say a vehicle is at a position $x_s$ at the start of a time epoch of $\Delta$ units. 
The idea is to exhaust these $\Delta$ units along the (prefixed) path of the vehicle and arrive at the 
appropriate end position $x_e$. Suppose $x_s$ is at a link $i$ and at a distance
of $d_e$ from the end of link $i$. Given that link $i$ is in state $s$, we sample once from $\mathcal{N}(y;\mu^{i,s}, 
\sigma^{i,s})$ to obtain a realization of $\tau^{i,t}$. Let $y$ be the
sampled value which corresponds to the time to travel from the start to the end of link $i$ in the current epoch. 
By linear scaling the time
taken to traverse a distance of $d_e$ would be $y\frac{d_e}{len(i)}$. If this time is greater than $\Delta$, then the vehicle 
doesn't cross link $i$ in
the current time epoch and settles into a position $x_e$ which is $len(i)\frac{\Delta}{y}$ units further from $x_s$. However if 
the time to traverse the
rest of the link, namely $(yd_e)/len(i)$, is less than $\Delta$, then the vehicle is moved to the start of the next link in its prefixed 
path. Further, now a
residual time of $\Delta - yd_e/len(i)$ needs to exhausted along the subsequent path  and the process described is continued 
till the residual
time is completely exhausted to reach a final position $x_e$. This process of trajectory formation is carried out for each vehicle 
at each discrete time epoch. 
\smallskip

\begin{figure}[t]
\center
\includegraphics[scale=0.7]{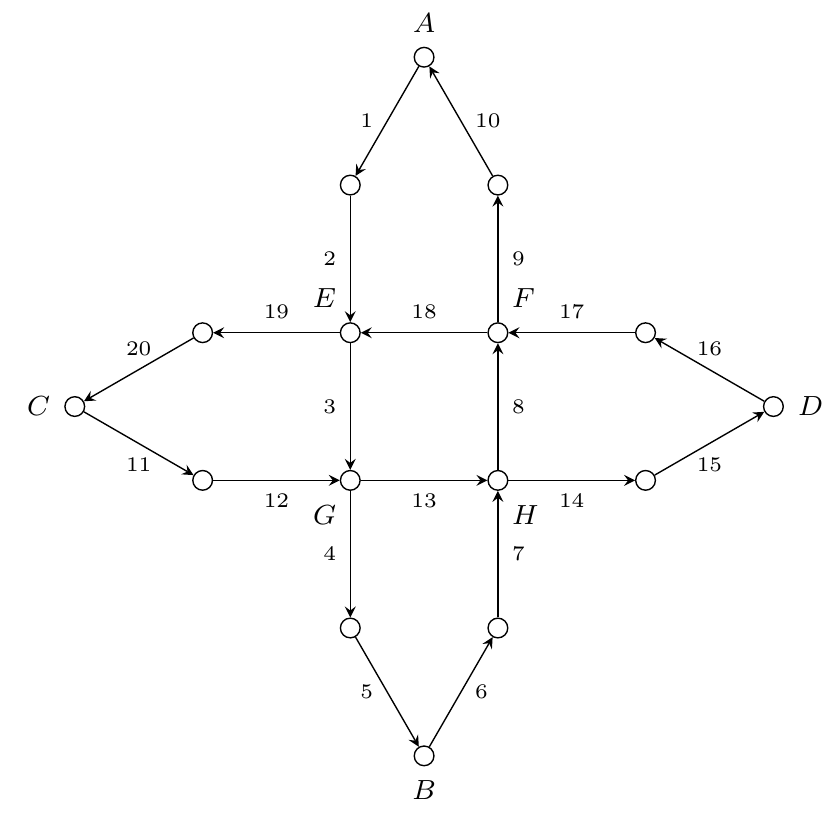}
\caption{Synthetic topology of 20 links}
\label{fig:10Link}
\end{figure}
\noindent
{\bf Network Structure:} The network structure chosen consists of 20 links as shown in figure~\ref{fig:10Link}. 
We chose this structure since it represents the gridded one-way roads that are typically found in the downtown 
area of various cities.
It consists of a long sequence of unidirectional lanes in either direction connecting point $A$ (in the north) to a point $B$ (in the south).
Traffic moving from $A$ to $B$ (north to south) have to go along links $1$ to $5$, while travelers need to take links $6$ to $10$ for
their  return.  Similarly there is a road infrastructure that connects $C$ (in the east) to $D$ (in the west) via links $11$ to $20$. 
This synthetic topology has a relatively 
rich neighbourhood structure with peripheral links like $1$, $6$, $11$, $16$ having $3$ neighbors (including themselves) while links $3$, $13$, $8$ and $18$ at the center 
having up to $5$ neighbors (including themselves). 
All links are assumed to have the same length of $1$ km.  For each link $i$, $\mu^{i,0} = 1.5$ min, $\mu^{i,1} = 3.0$
min,  
$(\sigma^{i,0})=(\sigma^{i,1})=0.1$. With $\Delta$ fixed at $5$ minutes, we chose about $60$ time steps for data generation and generated such sequences for $8$ days ($8$ i.i.d.
realizations). We choose to use $8$ probe vehicles which move along the north-south circular loop between A and B. Similarly, we also have another set of 8 probe vehicles moving east-west between C and D.
\smallskip

\noindent
{\bf Congestion Patterns:} 
We embed a short-lived congestion which can randomly originate at either link $1$, 
$6$, $11$ or $16$. Once congestion starts at a link, say link $1$, it moves downstream to link $2$ with probability $1$ at the next time step and this process continues
unidirectionally till link $5$. A similar congestion pattern which moves downstream one link at a time at every subsequent time step is embedded starting from link $6$, 
$11$ and $16$. Congestion doesn't persist in the same link into the next time step in any of the links -- these are short-lived congestions.  
Such short-lived congestion happens in real-world when a wave of vehicles traverse the links.
 
The above described short-lived congestion can be modeled using a NoisyOR based data generator.
The random chance of  congestion originating at link $1$ can be captured by setting $p_{1,0}$ to a low value, say $0.2$. 
The rest of the neighbors
of $1$, (namely $1$, $2$ and $10$) do not influence $1$ and hence the respective $p_{1,j}$s will all be zero. At link $2$,  to embed the
$1^{st}$ link's strong influence,  we set $p_{2,1}=1$. Since there is no congestion persistence, we set $p_{2,2}=0$. Since
we assume there is no upstream influence of congestion, we set  $p_{2,3}=0$. Similarly, the parameters for other
links are chosen.

We also  generate data with non-zero probabilities for congestion to persist for longer duration in a link. 
This means for a link $i$, $p_{i,j}$ (where $j$ refers to the self-neighbor) must be set to a non-zero
value.  We have chosen this value to be around $0.1$ across all links. This non-zero chance ensures some of the links once congested will continue to remain congested for one
or more subsequent time steps. 
\section{Map of the considered regions}
\vspace{-0.11in}
\label{sec:Porto}
\begin{figure}[h]
\center
\includegraphics[scale=0.38]{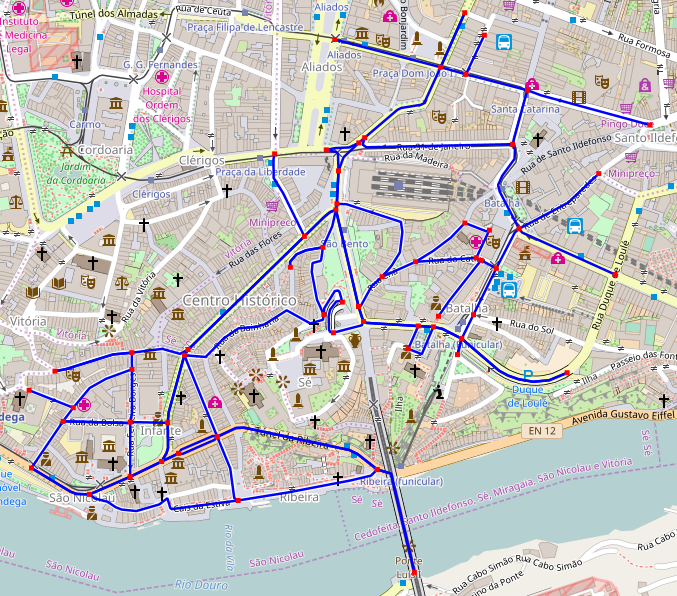}
\caption{Region of Porto on which Cab traces were considered.}
\label{fig:Porto}
\end{figure}
\begin{figure*}[t]
\center
\includegraphics[scale=0.48]{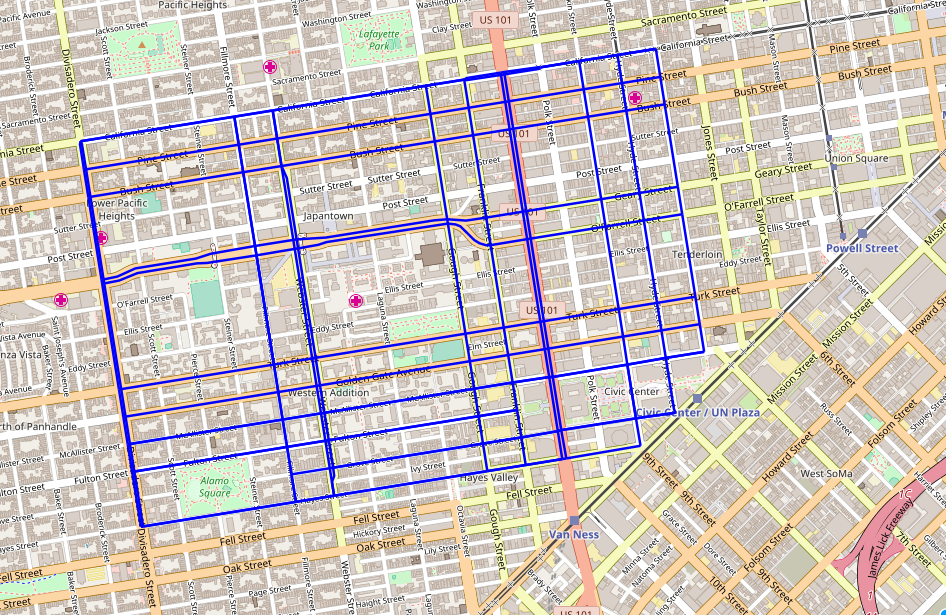}
\caption{Region of SanFransisco on which Cab traces were considered.}
\label{fig:SFO}
\end{figure*}


\end{document}